\documentclass[manuscript]{acmart}
\usepackage{algpseudocode}
\usepackage{algorithm}
\usepackage{caption}
\usepackage{subcaption}

\AtBeginDocument{%
  }
\usepackage{multicol}
\setcopyright{acmcopyright}
\copyrightyear{2022}
\acmYear{2022}
\acmDOI{XXXXXXX.XXXXXXX}

\begin{document}

\title{Mitigating Human and Computer Opinion Fraud via Contrastive Learning}

\author{Yuliya Tukmacheva}
\affiliation{%
  \institution{Skolkovo Institute of Science and 
Technology}
  \postcode{121205}
  \city{Moscow}
  \country{Russia}
}
\author{Ivan Oseledets}
\affiliation{%
  \institution{Skolkovo Institute of Science and 
Technology}
  \postcode{121205}
  \city{Moscow}
  \country{Russia}
}
\additionalaffiliation{%
  \institution{Artificial Intelligence Research Institute}
  \postcode{105064}
  \city{Moscow}
  \country{Russia}
}

\author{Evgeny Frolov}
\affiliation{%
  \institution{Skolkovo Institute of Science and 
Technology}
  \postcode{121205}
  \city{Moscow}
  \country{Russia}
}
\email{evgeny.frolov@skoltech.ru}

\renewcommand{\shortauthors}{Tukmacheva et al.}

\begin{abstract}
We introduce the novel approach towards fake text reviews detection in collaborative filtering recommender systems.
The existing algorithms concentrate on detecting the fake reviews, generated by language models and ignore the texts, written by dishonest users, mostly for monetary gains.
We propose the contrastive learning-based architecture, which utilizes the user demographic characteristics, along with the text reviews, as the additional evidence against fakes.
This way, we are able to account for two different types of fake reviews spamming and make the recommendation system more robust to biased reviews.
\end{abstract}

\maketitle

\section{Introduction}

The recommender systems are deeply integrated into many popular online commercial platforms at present.
These platforms collect data about both users' and items' attributes, as well as accumulate the ratings and feedback of products and services, to develop algorithms for significant enhancement of users' experience on the marketplace.
These algorithms are capable of influencing the purchasing behavior of users by ($1$) offering them the selection of the most relevant personalized positions, ($2$) reducing the individual searching costs, and ($3$) alleviating the information asymmetry on large commercial platforms with homogeneous sellers and products through feedback mechanisms.  
Since recommender systems have the power to affect the marketing decisions of users, they have become an attractive target for ratings and reviews manipulations, also known as attacks.
Specifically, these attacks are aimed at inflating/deflating the ranks and text reviews of certain product positions or at simply sabotaging the efficiency and credibility of the the commercial platform in general.

The current study focuses on solving the task of filtering out the deceptive opinions and detecting anomalous behavior on a platform with text reviews.
The emphasis on text reviews can be explained by the fact that texts are a more informative and a more reliable source of product's and seller's quality, than a star-rating system, which is easy to manipulate (see \cite{luca2011}, \cite{iltaf2013}, \cite{yang2018}, \cite{zhou2018}).
The deceptive opinions can be generated in two ways: ($1$) by hiring the professional human writers to review the product for monetary gains, and ($2$) by exploitation of state-of-the-art (SOTA, here and after) automated language models to imitate the human speech, and this distinction is crucial, since the existing algorithms do not address this difference and focus merely on filtering out the artificial text reviews \cite{salminen2022} and do not account for the problem of hiring the professional writers.
One should not underestimate the size of the dishonest reviews writers market, since almost $4\%$ of all online reviews are the dishonest opinions of real humans, and this leads to over $\$152$ billion costs in revenues of online marketplaces\footnote{https://theprint.in/opinion/almost-4-of-all-online-reviews-are-fake-their-impact-is-costing-us-152-billion/715689/.}\footnote{https://www.cs.uic.edu/~liub/FBS/fake-reviews.html.}.

The purpose of the current study is to develop the machine classifier, which the recommender platforms can use to: ($1$) detect opinion fraud attacks, which are generated by SOTA generative language models, such as the transformer-based Generative Pre-trained Transformer (GPT)-models (see \cite{radford2019language}, \cite{brown2020language}), and ($2$) filter out the biased texts of dishonest human writers, and, hence, ensure the system's robustness to these two type of attack strategies.
Generally, the recommender systems, especially, the collaborative filtering-based\footnote{The collaborative-filtering-based recommender systems exploit the connections in clusters of users with similar preferences to make automated recommendations.} ones, face the problem of the cold-start, such that the new users, who do not have the history of interactions, behavioral traces or defined preferences, cannot be automatically and adequately distinguished from the newly created profiles.
The current study covers the limitations of previous works and addresses both types of attacks in a cold-start regime.
To achieve this, we exploit the user demographic characteristics, alongside the text reviews, as an additional source of evidence against fake opinions \cite{salminen2022}, following the premises of paper of \cite{aktumak2019}, which uses the user attributes to detect the attack on a star-rating system.
The logic behind this is the following: the users with the same attributes are most likely to have the same preferences.
Since the demographic information about users is not a common knowledge and is available to the system's administrator(s) only, the attackers would create the user profiles with randomly assigned metadata.
This way we assume that the genuine text reviews are attached to user attributes in latently, so that the fake texts (both written by humans and language models) do not demonstrate such a connection.

The main contributions of the paper are as follows:
\begin{itemize}
\item The setup of connecting the text reviews with user metadata to filter out the fake text reviews is new and has not yet been covered by existing works. 
The idea of exploiting the latent connections between the text reviews and demographic metadata has not been tested yet, and the recommender systems may use the proposed approach to filter out the fake text reviews more efficienly than before.
Thus, the present research becomes the baseline for other studies in the following area.
\item The current work addresses the problem of detecting both the dishonest users and the usage of generative language models.
The existing algorithms and approaches concentrate exclusively on the detection of artificially generated texts and ignore the cases of dishonest humans writing the biased texts for monetary gains.
The proposed method accounts for these two types of attacks, making the recommender system more robust and trustworthy.
\item The present research enriches the data base for fraud detection in recommender systems with two data sets.
The first data set is a processed and enlarged Google Local Business Reviews and Metadata, extended with carefully engineered features obtained from user metadata and supplemented with labels, extracted from proposed sampling procedure.
This data set can further be used to train and test the algorithms for real users' dishonest reviews.
The second data set is the collection of fake text reviews, generated by the GPT-J-6B model \cite{gpt-j} and augmented with randomly assigned user demographic attributes, which can further be exploited for training and testing of the models for artificially generated text reviews detection.
The processing and engineering of these defined data sets is described in more details in \hyperlink{ch:3}{Section 3} and \hyperlink{ch:4}{Section 3}.
\end{itemize}

The rest of the paper is organized as follows.
\hyperlink{ch:2}{Section 2} covers the existing papers on fraud detection in recommender systems and discusses their shortcomings.
\hyperlink{ch:3}{Section 3} describes the methods and introduces the theoretical framework of the current research.
\hyperlink{ch:4}{Section 4} covers the empirical application, reports the
estimation results and runs the comparison experiments.
\hyperlink{ch:5}{Section 5} discusses the obtained results, reviews the research limitations and concludes.

\section{Related Work}\hypertarget{ch:2}{}
The industrial need for robust and effective fake detection algorithms prompts the extensive research in this study area.
One promising direction of the research in attacks detection on recommender systems is devoted to simulation of attacks in a dynamic Nash reinforcement learning setting (see \cite{dou2020}, \cite{lin2020}, \cite{tang2020}).
This approach implies the simulation of the attack environment and defining the cost functions of malicious users to deduce their optimal strategy and to counteract it.
However, this method is not valid in the setting of large online marketplaces in which the heterogeneous attackers with different cost functions and motives may be present.
The proposed approach covers this limitation and is capable of working with heterogeneous malicious agents and does not depend on attackers' strategy approximation.
Also, the proposed approach is able to cope with the setup in which the same user may be defined as malicious and reliable, when the real human acts as a spammer for some category of items and as an ordinary customer for another category.
Our model considers each pair of attributes and texts as independent of the previous interactions and is able to account for such (although, practically rare) situations.

Another group of models for fraudulent profiles detection exploits diverse collection of statistical metrics for filtering the anomalies.
Existing works propose such statistics as ($1$) deviation from the median value in a suggested recommendation set of products \cite{iltaf2013}, ($2$) rating items with extreme rating and sentiment values \cite{yang2018} and ($3$) ratings falling within the tight time windows \cite{zhou2018}.
Nonetheless, the proposed statistics are easy to circumvent and the more advanced detection metrics should be applied.

There is another direction of research, which utilizes the clustering methods (see \cite{lee2012}, \cite{bilge2014}).
These models try to determine the highly-connected cliques of users and exploit the clustering methods to determine anomalous profiles.
The SOTA clustering models for fakes filtration are Support Vector Machine-based (SVM) models. In \cite{alostad2019} the Support Vector Machine + Gaussian Mixture Model (SVM-GMM) is introduced, and in \cite{samaiya2019} the Support Vector Machine + Principal Component Analysis (SVM-PCA) approach is used.
However, the named approaches suffer from the cold start problem.
The proposed approach accounts for each observation as a new one and, thus, detects the fakes in a cold start regime. 

More advanced techniques for attack detection in recommender systems are the latent variable models, which encode the genuine and fake types of users.
In \cite{li2011}, the latent variable model is employed to reveal the user type by solving the program of maximization the entropy of ratings distribution in a star-rating system.
In \cite{wu2021}, the latent variable model relies on counteracting the adversarial poisoning attacks, which is the process of fake profiles generation that is intended to minimize the empirical risk.
The algorithm described in \cite{aktumak2019} also uses the generative factor model, which utilizes the user metadata as an additional source of evidence against fake ratings.
We use the premises of this model in the current study, though all of the named variational inference models are developed for star-rating platforms and ignore the text reviews manipulation.
As outlined above, the star-rating system on its own is not very informative for users, and the text reviews are considered to be a more reliable source of information for customers.
Therefore, the approach which is capable of working with text reviews, written both by humans and language models, is more socially desirable.
Thus, the assumptions mentioned in \cite{aktumak2019} has inspired the usage of user metadata to generate the model, capable of working with text reviews and of filtering two types of attack strategies.
This problem has not yet been investigated by the existing papers, and, henceforth, there are no baselines to compare the proposed approach with.
The SOTA approach for determining the fakeness of the text review based on the review only has been proposed in \cite{salminen2022}.
It introduces the procedure of fake reviews generation, using the GPT-2 model \cite{radford2019language}, and the fine-tuning of a Robustly Optimized BERT Pretraining Approach (RoBERTa) \cite{DBLP:journals/corr/abs-1907-11692} model to classify the text reviews into real and fake.
We use this model as the benchmark to compare our proposed approach with.
This way we can also check if user metadata indeed helps to determine the fakeness of the review.

\section{Methods}\hypertarget{ch:3}{}
\subsection{Fake Reviews Generation Procedure}

\subsubsection{GPT-J-6B model}
To simulate the fake reviews attacks, we intend to use the SOTA unsupervised transformer-based multitask language models, specifically, the Generative  Pre-trained Transformer (GPT here and after) models (see \cite{radford2019language}, \cite{brown2020language}).

These models are supposed to perform multitask learning, such as machine translation, text summarization and comprehension, as well as question answering.
The current research exploits the GPT-J-6B \cite{gpt-j} model, which is a democratized open-source version of the GPT-3 model \cite{brown2020language}
, latest generation of the GPT-model, to generate the answer, following the processed query-answer history. 

GPT-J-6B model is an autoregressive model, produced by EleutherAI, for text generation, which is the stacked set of transformer decoder blocks, each of which is the sequence of masked self-attention layers, encoder-decoder self-attention layers and feed-forward neural networks.
It has 6 billion parameters and has been trained on the large Pile data collection \cite{DBLP:journals/corr/abs-2101-00027}, which is an $800$GB text data set for training the language models.
Given the text context, the model aims at predicting the next word in a given sequence.
Having been taught on a large scope of text data, the model demonstrates relatively good performance on different zero-shot down-streaming tasks \cite{floridi2020}\cite{budzianowski2019hello}, generating the syntactically concise texts, capable of mimicking the real human speech.
GPT-J-6B model relies on transfer learning, so that no additional training and data set adjustments are necessary to solve the task.
Therefore, we can form the query-answer prompts to be processed by the GPT-J-6B model to obtain the fake reviews on a large scale.

\subsubsection{Generation Procedure}
Having the collection of the text reviews as the examples of real texts, the malicious users can exploit the SOTA generative models to simulate the human written speech.
We have developed the procedure to generate fake reviews, based on the real reviews, with the following steps:
\begin{enumerate}
    \item Standard preprocessing of the text reviews;
    \item Extracting the keywords of each text review with TextRank4Keywords algorithm;
    \item Choosing a subset of keywords to send to prompt;
    \item Randomly changing the keywords to synonyms/antonyms;
    \item Forming the query-answer prompt to use in GPT-J-6B model for fake review generation;
    \item Sampling the user attributes and randomly match them with fake texts to generate the attackers' profiles data set;
    \item Merging the fake data set with a random subset of the original data set to get the balanced data collection for the fakes classification task.
\end{enumerate}

\paragraph{Step $1$} Standard preprocessing of the text reviews, such that the removal of redundant symbols and stopwords is performed.

\paragraph{Step $2$} Then we should shuffle the data set and group it by categories.
For each category, iteratively, we take $4$ reviews and extract the keywords from each of them with TextRank4Keywords algorithm.
The TextRank4Keywords algorithm is based on PageRank algorithm calculates the weights for websites, representing the web pages as a directed graph, via the following formula:
\[
S(V_i) = (1-d) + d \cdot \sum_{j \in \mathrm{In}(v_i)} \frac{1}{|\mathrm{Out(V_j)}|} S(V_j),
\]
where 
$S(V_i)$ is the weight of $i$th node (web page), $d$ is the damping factor used for nodes with no outgoing connections (we set $d=0.85$), $\mathrm{In}(V_i)$ is a set of in-going connection of node $i$, $\mathrm{Out}(V_j)$ is a set of out-going connections of node $j$.

To use PageRank for texts, we split each review into sentences and specify the window size $k$ (we set $k=4$).
Thus, any two-word pairs in a window represent the undirected edge and then calculate the weights of such a graph, using the formula above.
After this, we take $10$ most important words in a text review.
For simplicity, we specify the POS tags of words and analyze only the nouns, verbs, adverbs and adjectives.

\paragraph{Step $3$} We randomly choose one-fourth of a set of unique keywords from each of $4$ processed text reviews in a category.
We will use it as basis words for prompt generation.

\paragraph{Step $4$} Since the GPT-based models seem to be sensitive to words used in previous queries, such that for seen words it returns the example answer, we should change all the words to their random synonyms.
Therefore, we change all words in the subset to their synonyms, if such exist.
Also, it is important to note that the attackers intend to inflate/deflate the rankings of products, so that they would generate the texts with desired sentiment.
For this reason, with probability $0.5$, we change each of selected key words to their antonyms to account for the case in which all of the reviews are majorly positive/negative.
This way, we would get the reviews with diverse sentiment and enrich the data collection with different texts.

\paragraph{Step $5$} We generate the query-answer prompt in the following way.
We take keywords from each review and consider them as queries.
The answers to these queries would be the real text review, corresponding to these keywords.
Then, as a query for the fake review, we send the random subset with synonyms/antonyms and ask the generative transformer-based model to generate the review.

\paragraph{Step $6$} Having the collection of fake review texts, one can match these texts with random user attributes.
Since the information about the distribution of user metadata is not a common knowledge, the attackers just randomly assign the user characteristics.
For this, we randomly sample the user attributes from real metadata and assign them to fake text reviews.

\paragraph{Step $7$} We choose the random subset of the original data set and merge it with the created data set.
This way, we get the balanced data set, twice as long as the fakes data set, which one can use for fakes classification task.
 
\subsection{Sampling Procedure for Human Spammers Detection}
Having the data set with real text reviews and corresponding user demographic characteristics, we can modify this data collection to simulate the human spammers attack.
To create the data set for fakes filtration amongst dishonest real users, one should perform the following pipeline, described in \hyperlink{alg:1}{Algorithm 1}.
Before sampling, one should obtain the embeddings of each text review in the data set, using the multilingual Bidirectional Encoder Representations from Transformers (BERT here and after) (cased) \cite{DBLP:journals/corr/abs-1810-04805}.
Then, for each text review we calculate the average text embedding of size $(1, 768)$.
Then, one should choose the random observation $k$ with text review and attributes from User $i$.
With probability $2/3$\footnote{The choice of probabilities ensures that, on average, the resulting data set would be balanced.} we leave the text review-attributes pair as is and label such pairs as positive. 
With probability $1/3$ we choose any User $j$ ($j \neq i$) to match the $k$th observation with. 
Then, among all the reviews written by User $j$, we take the one, which is the most dissimilar to 
$k$th text review (in terms of embeddings comparison).
For this, we calculate the dot product between the embedding vector of $k$th review and the matrix $Rev\_j$ of embedding vectors of all reviews written by User $j$.
After this we choose the text review $m$ with maximal dot product and swap the user attributes, such that for the $k$th text review we assign User $j$'s metadata and for the $m$th text review we assign User $i$'s metadata.
Such shuffled pairs we label as negative.
This way, we get the balanced data set with half of the user attributes shuffled to ensure that the connection between the text review and user demographic characteristics is lost (negative pairs) and half of the data left as is (positive pairs).
Then we should train the contrastive learning-based model to distinguish between the positive and negative pairs of user attributes embeddings and text embeddings, so that the embeddings for the negative pairs are maximally different and for the positive pairs are maximally close in terms of the Euclidean distance.
The contrastive learning setup for the current task is describes in more details in \hyperlink{ch:3}{Chapter 2, Section 2.3}.

\begin{algorithm}[H]
\caption{Sampling Procedure for Human Spammers Detection Data Set}\hypertarget{alg:1}{}
\begin{algorithmic}[1]
\Require shuffled data set with text review embeddings $\mathcal{D}$.
\Ensure balanced data set with negative and positive pairs.
\State $unprocessed\_idx \gets range (0, \text{len}(\mathcal{D}))$
\For {$k \gets 0$ to len($\mathcal{D}$)}
\If{$k$ in $unprocessed\_idx$}
\State $coin\_flip \sim \mathrm{Bern}(p=2/3)$
\If {$coin\_flip = 1$}
\State remove $k$ from $unprocessed\_idx$
\State add label $= 0$ to $k$th row in $\mathcal{D}$
\ElsIf{$coin\_flip = 0$}
    \State $user\_i \gets \mathcal{D}.username[k]$
    \State $rev\_ik \gets \mathcal{D}.review[k]$
    \State sample $user\_j $ from $\mathcal{D}.usernames \setminus user\_i$
    \State $Rev\_j = [\mathcal{D}.reviews[\mathcal{D}.username == user\_j][$idx if idx in $unprocessed\_idx]]$
    \State $max\_emb\_idx = \text{argmax} \left(Rev\_j \times rev\_ik \right)$
    \State swap text embeddings with indices $k$ and $max\_emb\_idx$
    \State remove $k$ from $unprocessed\_idx$
    \State remove $max\_emb\_idx$ from $unprocessed\_idx$
    \State add label $= 1$ to $k$th row in $\mathcal{D}$
    \State add label $= 1$ to $max\_emb\_idx$th row in $\mathcal{D}$
\EndIf
\EndIf
\EndFor
\end{algorithmic}
\end{algorithm}
\subsubsection{Multilingual BERT (cased)}
For obtaining the embeddings of the review texts, we use the multilingual BERT (cased) \cite{DBLP:journals/corr/abs-1810-04805}, since the data set contains text reviews written in different languages.

This transformer model is self-supervised and has been trained on the large Wikipedia collection of texts in $104$ languages with the masked language modeling objective.
The architecture of BERT is similar to GPT-based models and represents the stacked transformer encoder blocks, consisting of the sequence of self-attention layers and feed-forward neural networks.
This approach implies that the model randomly masks some words in the sentence (replaces them to the MASK token) and is trained to predict these masked words, and this ensures that the model gives the bidirectional text representation as the output.
We use the pre-trained multilingual BERT (cased) to extract the contextualized representations of each token (vector embeddings) in the text review.
First, we tokenize the sentences in each review, converting it to a list of words in the review and padding it to the maximal length or $512$, and pruning if the sentence is of length greater than $512$.
Then, we obtain the indices from the tokenizer output, so that for each token we get its index in the BERT vocabulary dictionary, and obtain the attention mask, which contains $1$s for unpadded tokens and $0$s for padded ones.
Next, we send the indices and attention mask to the BERT model to get the contextualized embeddings for each sentence in the text review.
After this, for each text review we compute the average text embedding vector to use in the contrastive learning setup.
\subsection{Contrastive Learning Setup}
We follow the approach described in \cite{1640964}.
Having the similar (positive) and dissimilar (negative) pairs of vectors as outputs of the $G_W$ deep neural network, parametrized by $W$, $G_W(X_1)$ and $G_W(X_2)$, we define the interpoint distance between them as the Euclidean distance (approximation of the ``semantic similarity''):
\[
D_W(X_1, X_2) = \left|\left|G_W(X_1) - G_W(X_2)\right|\right|_2.
\]
Then, we define the contrastive loss in the following way:
\begin{align*} 
\mathcal{L}(W) &= \sum_{i=1}^P L\left(\left(W, (Y, X_1, X_2\right)^i\right),\\
L\left(W, \left(Y, X_1, X_2\right)^i\right) &= (1-Y) L_S \left(D_W^i \left(X_1, X_2\right)\right) + Y L_D \left(D_W^i\left(X_1, X_2\right)\right),
\end{align*} 
where $X_1$ and $X_2$ are the vectors in pair, $Y = 0$ is a target label for similar (positive) pairs, $Y = 1$ is a target label for dissimilar (negative) pairs, $\left(Y, X_1, X_2\right)^i$ is the $i$th sample pair, $L_S$ is the partial loss function for positive pairs, $L_D$ is the partial loss function for negative pairs, and $P$ is the number of samples.

One should use such functions $L_S$ and $L_D$, such that minimization of $L$ with respect to $W$ results in low values of distance $D_W$ for positive pairs and high values of $D_W$ for negative pairs.
The choice of the corresponding partial loss functions satisfies this condition:
\begin{align*}
L_S \left(W, X_1, X_2\right) = \frac{1}{2} \left(D_W \left(X_1, X_2\right)\right)^2 \\
L_D \left(W, X_1, X_2\right) = \frac{1}{2} \left(\max\left(0, m-D_W \left(X_1, X_2\right)\right)\right)^2,
\end{align*}
where $m > 0$ is the margin, which corresponds to the radius around $G_W(\cdot)$, so that the negative pairs are accounted for only if they are inside the margin.
This can be seen through computation of the gradients of these functions:
\begin{align*}
\frac{\partial L_S}{\partial W} &= D_W \frac{\partial D_W}{\partial W} \\
\frac{\partial L_D}{\partial W} &= 
\begin{cases}
0, \text{ if } D_W > m,\\
-(m - D_W) \frac{\partial D_W}{\partial W}, \text{ otherwise},
\end{cases}
\end{align*}
so that low values of $D_W$ generate a gradient to decrease $D_W$ in case of $L_S$, and the reverse situation is valid for case of $L_D$ because of the negative sign.

The current approach solves the problem of minimization of the defined contrastive loss in the deep neural network setting, which is covered in the next subsection.

\subsubsection{Deep Neural Network Architecture}

We use the multilayer perceptron (MLP) architecture, similar to siamese ones, introduced in \cite{e6377ee676a34e8eb97cfdc53cd489ef}, \cite{10.5555/2987189.2987282}.
It consists of two branches of function $G$ with the same parameters set, corresponding to processing (1) the average text embedding vector ($X_1$) and (2) the embeddings of categorical and numerical features ($X_2$), as defined in \hyperlink{fi:architecture}{Figure 1}\footnote{Based on approach described in https://yashuseth.wordpress.com/2018/07/22/pytorch-neural-network-for-tabular-data-with-categorical-embeddings.}.
Both of the inputs ($X_1$ and $X_2$) are passed through the corresponding branches, yielding the outputs $G^1_W(X_1)$ from the first branch and $G^2_W(X_2)$ from the second branch, respectively.
Then, we calculate the pairwise distance between both outputs of the branches, which represent the vectors, and then use this distance scalar value to calculate the partial losses ($L_S$ and $L_D$) and the total contrastive loss $L$.
Both outputs of the branches represent the vectors and the label of the pair ($Y$) are then passed to the contrastive loss.
The parameters of the network $W$ are then updated via Adam optimization algorithm, so that the gradients are calculated through back-propagation of the contrastive loss value. 
The total gradient is the sum of contributions of two branches' outputs.

We also use the regularization tricks, such as the weights initialization, dropouts and batch normalizations, to tackle the overfitting.
The results of experiments with generated data sets are present in \hyperlink{ch:4}{Chapter 3}.

\begin{figure}[h!]
\includegraphics[width=\textwidth]{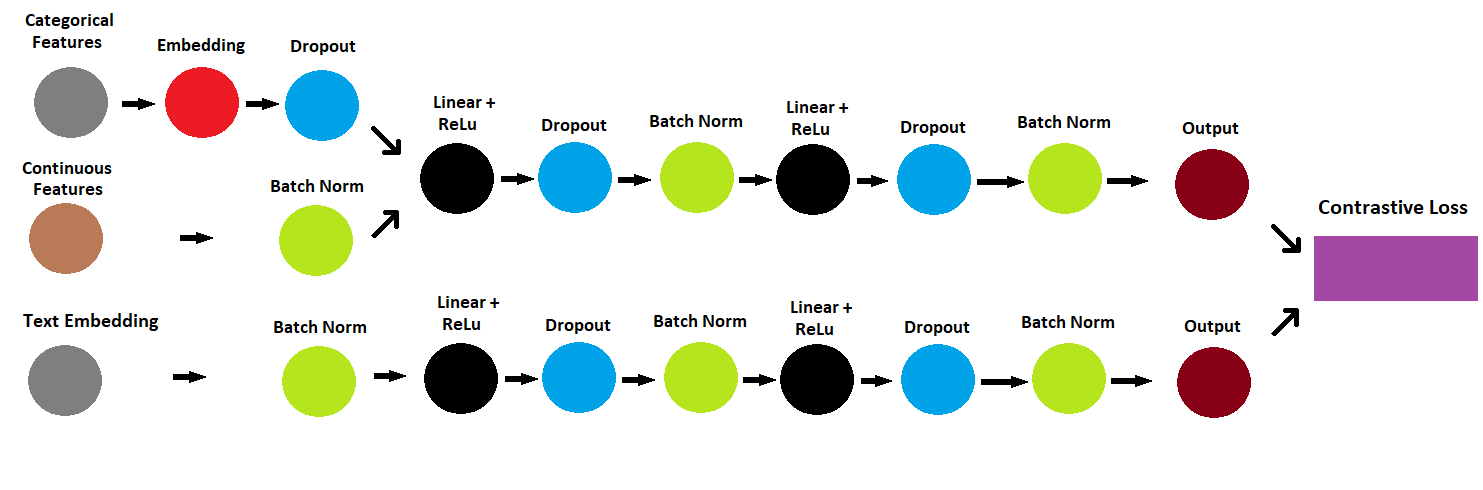}
\caption{Multilayer Perceptron (MLP) architecture, which has been used in the outlined contrastive loss setting.}
\hypertarget{fi:architecture}{}
\end{figure}

\subsection{fakeRoBERTa Benchmark}
As it has been stated above, the problem statement of the current research has not been investigated in the recommender systems community.
Therefore, there are no valid algorithms to compare our model with in terms of quality metrics.
The SOTA approach in the topic of fake reviews detection has been proposed in \cite{salminen2022} and creates the fakeRoBERTa language model to classify text collection into the fake (generated by GPT-2 \cite{DBLP:journals/corr/abs-1810-04805} model, discussed above) and the real ones.
We will use this approach as the benchmark and check if the exploitation of the user metadata helps in the fake reviews detection.

The authors collect the data set of fake reviews, generated by the GPT-2 model and then fine-tune the RoBERTa model for the specified classification task.
They use different architectures of language models to account for the possibility of data leakage and classification model's overfitting to dependent generative model's language style.
The results show that fakeRoBERTa has the accuracy of $96\%$ and it is the highest performance score amongst compared OpenAI models on text classification tasks.
Also, the authors conduct the crowdsourcing experiment regarding the comparison between human and computer detection of fake reviews.
They have found that human performance is insignificantly better than the random classification on the generated data set, implying the advances of attacks, employing the generative language models.
However, the authors address the problem of fake generative reviews detection, ignoring the human spammers, writing the reviews for monetary gains.

This chapter introduces the numerical experiments conducted in the current study.
First, it outlines the languages and programs that have been used in the research.
Then, it describes in detail the data set that has been used for training and validation of the proposed contrastive learning-based approach.
After this, it introduces the quality metrics that has been used for checking the performance of the proposed approach and its comparison with the SOTA RoBERTa benchmark for determining the genuineness of the text review.
Finally, the experiment results and the benefits of the proposed method are provided.

\section{Numerical Experiments} \hypertarget{ch:4}{}
\subsection{Data Sets}

\subsubsection{Data Set for Human Biased Reviews Filtration}
\paragraph{Data set description. } The data set that satisfies the setup of the current study should contain both users' demographic characteristic and the corresponding text reviews.
The only open-source data set available for training and testing the proposed approach is the Google Local Business Reviews and Metadata\footnote{https://cseweb.ucsd.edu/~jmcauley/datasets.html\#google\_local.} \cite{he2017} \cite{pasricha2018}.
This data set from Google is a large collection (approximately $11.5$ million) of text reviews and $5$-star ratings from $4.5$ million users on $3.1$ million local businesses ($48$k categories of restaurants, shops, schools, hotels, beauty salons, hospitals, etc) all around five continents.
The data set also contains customers' demographic characteristics, such as user names and ids, GPS-coordinates of the current location, coordinates of previous places lived, education and job history (places and positions/majors), and review timestamps.

We have carefully processed the defined data set and extended it with several engineered features.
The final augmented Google Local Business Reviews and Metadata has the following features:

\begin{multicols}{3}
\begin{enumerate}
\item current country (categorical: $200$ categories),
\item number of places lived (numerical),
\item education degree (categorical: $5$ categories),
\item number of places studied (numerical),
\item education major (categorical: $101$ categories),
\item current profession (categorical: $101$ categories),
\item number of previous jobs (numerical),
\item unemployment (categorical: $2$ categories),
\item retired (categorical: $2$ categories),
\item local business category (categorical: $51$ categories),
\item text review length (numerical),
\item number of unique words in the review (numerical).
\end{enumerate}
\end{multicols}

We have also performed the sampling procedure for the contrastive learning setup, described in \hyperlink{ch:3}{Chapter 2} and added the binary label target variable, which takes value $0$ if the text-attributes pair is positive and $1$, otherwise.
This way, we get half of the data marked as positive pairs, which assumes the latent relation between user metadata and review texts, and half of the data marked as negative pairs, such that the user attributes are shuffled and assigned to text reviews of other users in a non-random way.
Henceforth, we get the data set, which represents the attack of human spammers, who write biased text reviews.

\subsubsection{Data Set for Generative Language Models Texts Detection}

We have also compiled a new data collection of fake reviews, generated by language models.
We have formed the query-answer prompts for the GPT-J-6B model \cite{gpt-j}, based on Google Local Business Reviews and Metadata texts, following the procedure described in \hyperlink{ch:3}{Chapter 2}.
According to the premises of the \cite{aktumak2019}, we assume that the malicious users, who do not have access to the distribution of user attributes, would fill in the demographic information about themselves in a random way, and, therefore, the connection between the user attributes and text reviews would be violated in case of fake profiles.
Henceforth, to simulate the attack of fake profiles, we have randomly sampled the user attributes for each of the fake review text and have obtained the data set, which simulates the attack of generative language model with fake random user metadata.

\subsection{Metrics}
In the current research we intend to train the multilayer perceptron (MLP), defined in \hyperlink{ch:3}{Chapter 2}, to reduce the distance between outputs of both branches if the input pair has been positive, i.e. there is the match between the user demographic characteristics and the written text, and increase the distance, otherwise.
For this we solve the optimization problem of the contrastive loss minimization.

Since both of the generated data sets are balanced, we can use the accuracy as the quality metric.
Having the trained model, which pushes aside the attributes embeddings and average text embeddings for negative pairs and brings together these embeddings for positive pairs, during evaluation we can compare the pairwise distance between model's outputted vectors with the threshold.
For the current experiments we take the threshold equal to $0.5$.

The fakeRoBERTa benchmark also uses the accuracy as a quality metric.
It also minimizes the negative likelihood loss, which is defined as
\[
\log \mathbb{P}(\mathcal{D} | \theta) = \sum_{i=1}^n \left(y_i \log \hat{y}_{\theta, i} + (1-y_i) \log \left(1-\hat{y}_{\theta, i}\right) \right),
\]
where $\mathcal{D}$ is the data observed, $\theta$ denotes the parameters to optimize, and $\hat{y}_{\theta, i}$ is the probability of the $i$th observation to be labeled positive.

\subsection{Experiment Results and Comparison with Existing Algorithms}

We have compared the performance of the proposed algorithm in two classification settings: \begin{enumerate} \item distinguishing between real humans writing the text reviews with their user metadata unchanged (i.e. the connection between the user demographic characteristics and text review embeddings is left as is), and real humans writing biased reviews with their user metadata shuffled (i.e. the defined connection is violated), and \item distinguishing between the text reviews written by real honest humans and by generative language models.\end{enumerate}

\paragraph{Honest Users vs Human Spammers. }
We have used the proposed contrastive learning setup, on one of the generated data sets (based on the Google Local Business Reviews and Metadata, consisting of $496'320$ observations) with half of the data left as is (positive pairs) and half of the data shuffled (negative pairs), according to the procedure, defined in \hyperlink{ch:3}{Chapter 2}.
We split data into $80$-$20$\% chunks for training and validation sets.
We have trained the proposed architecture for $30$ epochs.
The results are shown in \hyperlink{ta:res1}{Table 1} and the training and validation metrics behavior is given on \hyperlink{fi:res1b}{Figure 2}.

\begin{table*}[!htb]
    \centering
     \caption{
     Loss and Accuracy values for training and validation sets on Honest Humans vs Human Spammers and Honest Humans vs Language Model classification tasks for the proposed approach and the fakeRoBERTA benchmark.}
    \begin{tabular}{p{3.5cm}p{3.5cm}p{3.5cm}p{3.5cm}}
    \toprule
        Setup & Humans vs Humans (Contrastive Learning) & Humans vs Language Model (Contrastive learning) & Humans vs Language Model (fakeRoBERTA)\\
        \midrule
      Training Contrastive Loss   & $1213.82$ & $22393.25$ & -\\
      Validation Contrastive Loss   & $1112.41$ & $22466.42$ & - \\
      Training NLLoss & - & - & $1.01$\\
      Validation NLLoss & - & - & $0.96$ \\
      Training Accuracy & $0.73$ & $0.62$ & $0.49$\\
      Validation Accuracy & $0.66$ & $0.60$ & $0.5$\\
      \bottomrule
    \end{tabular}
    \hypertarget{ta:res1}{}
\end{table*}


\begin{figure}
\hfill
\begin{subfigure}[t]{0.45\textwidth}
    \includegraphics[width=\textwidth]{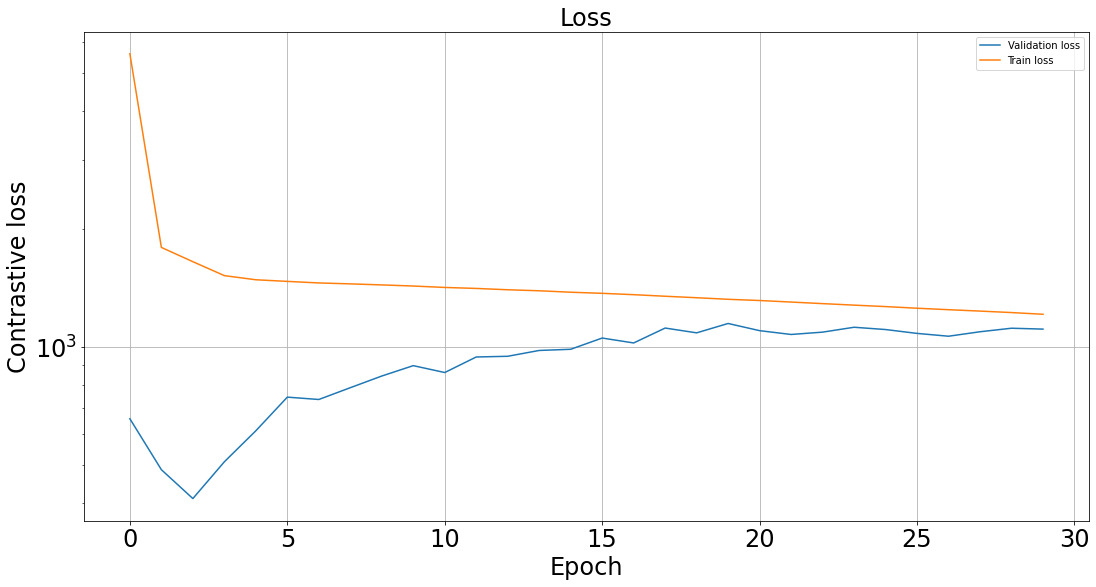}
\end{subfigure}
\hfill
\begin{subfigure}[t]{0.45\textwidth}
    \includegraphics[width=\textwidth]{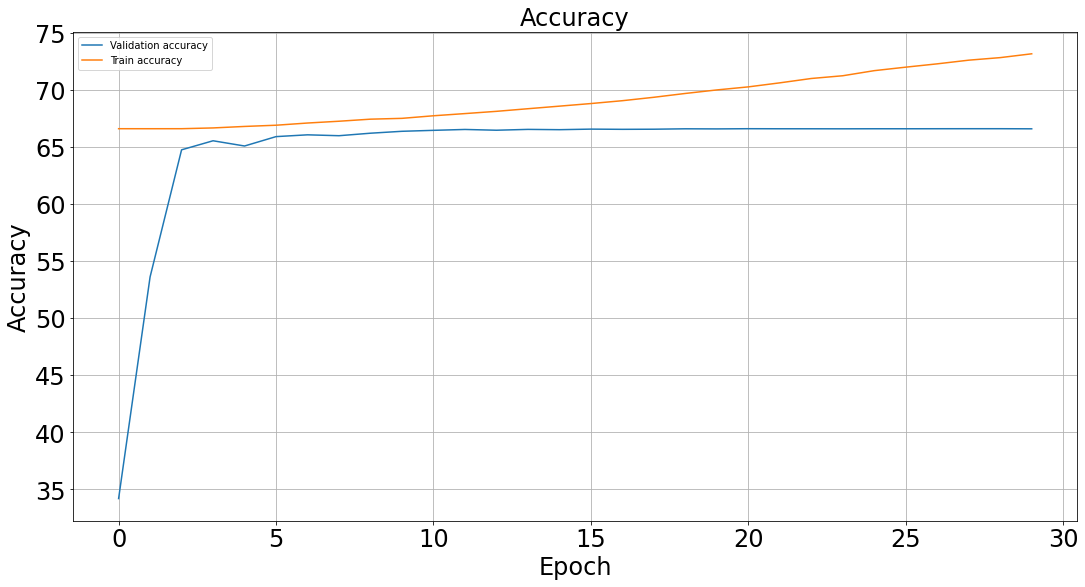}
\end{subfigure}
\hfill
\caption{Contrastive Loss and Accuracy metrics plots for training (orange) and validation (blue) sets on Honest Humans vs Human Spammers classification task.}
\hypertarget{fi:res1a}{}
\hypertarget{fi:res1b}{}
\end{figure}

The accuracy metrics are not very high and display not very stable training performance, but they set a good baseline for future algorithms to be built upon.
These values also set a good research question: whether there is a strong connection between the texts, written by people, and their demographic attributes.
It may be the case that this connection is too subtle to be grasped by machine models.

\paragraph{Honest Users vs Language Models. }
We have also utilized the proposed approach for another generated data set, which consists of $14k$ real text reviews and user attributes (positive pairs), randomly chosen from the original Google data set, and $14k$ artificially generated text reviews with randomly assigned user metadata.
We have also split the data into $80$-$20$\% training and validation chunks, respectively, and have also trained the model for $30$ epochs.
The values of the contrastive losses and accuracy metrics for training and validation data sets can be found in \hyperlink{ta:res1}{Table 1} and on \hyperlink{fi:res2a}{Figure 3}.

\begin{figure}
\hfill
\begin{subfigure}[t]{0.45\textwidth}
    \includegraphics[width=\textwidth]{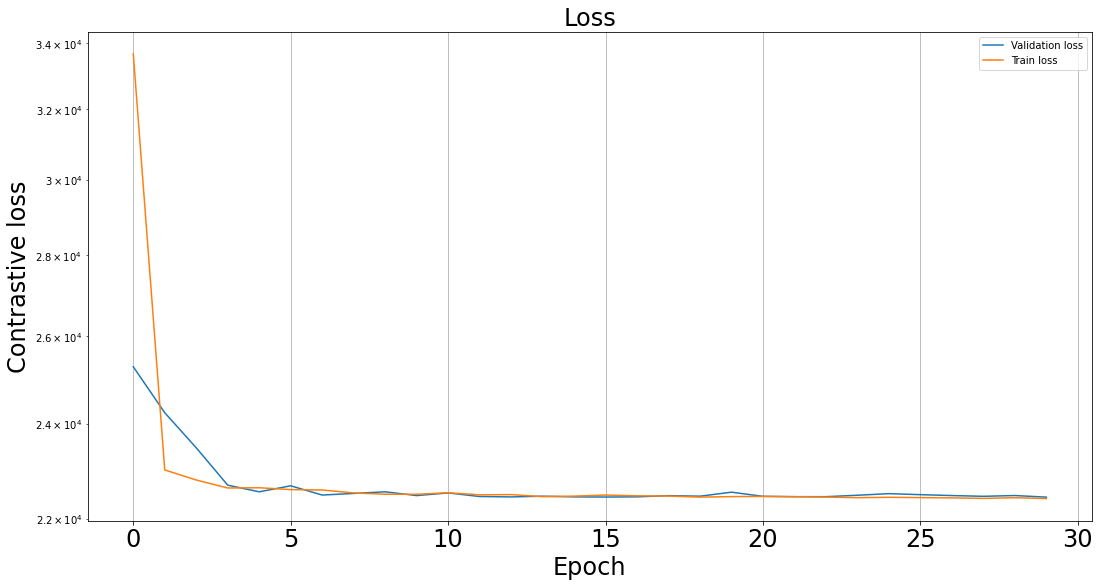}
\end{subfigure}
\hfill
\begin{subfigure}[t]{0.45\textwidth}
    \includegraphics[width=\textwidth]{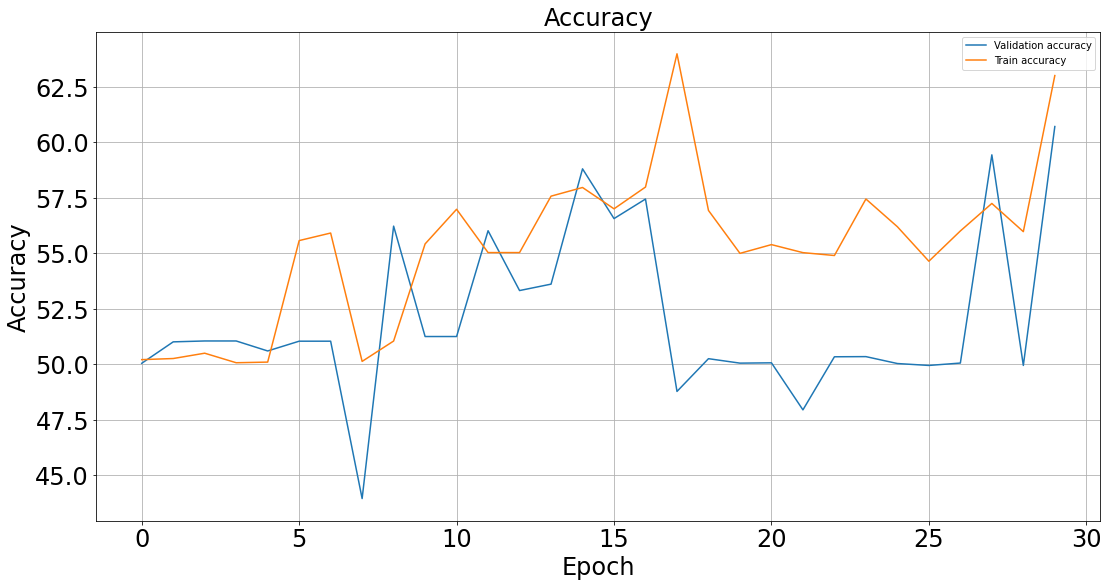}
\end{subfigure}
\hfill
\caption{Contrastive Loss and Accuracy metrics plots for training (orange) and validation (blue) sets on Honest Humans vs Language Model classification task.}
\hypertarget{fi:res2a}{}
\hypertarget{fi:res2b}{}
\end{figure}

One can observe that both metrics are worse off, compared to Humans vs Humans setup.
It may be explained by the fact that the language model has been trained on real reviews and may have given overfitted text reviews as outputs and the fact that the training data set is not large enough (only $28k$ observations).

\paragraph{Comparison with RoBERTa. } As it has been outlined before, the approach, proposed in the current study, is novel and has not any valid benchmarks to compare with.
Nonetheless, we are able to conduct comparison of the proposed algorithm with the SOTA method for fake reviews detection is the fakeRoBERTa \cite{salminen2022}, which has been trained to distinguish real reviews from the ones generated by GPT-2 model \cite{DBLP:journals/corr/abs-1810-04805}.
We have followed the training pipeline, provided by authors, and have fine-tuned RoBERTa to classify real reviews and the ones, generated by GPT-J-6B \cite{gpt-j} on $5$ epochs.
The accuracy and negative likelihood loss values are shown in \hyperlink{ta:res2}{Table 1}.

As it can be observed from training results, the performance of fine-tuned fakeRoBERTa is no better than random guess, so that the generated reviews are very similar to the real ones.
It also shows, in comparison to proposed approach, that the user demographic attributes, indeed, help to distinguish between real and generated text reviews.

\paragraph{Benefits of the Proposed Method. }
The experimental results show the significant improvement of the proposed method over the SOTA fakeRoBERTa for fake reviews detection in terms of the accuracy metric.
Also, the studied algorithm shows decent quality for both classification tasks, making it efficient for fakes filtration among humans and language models.
Nonetheless, a deeper linguistic and psychological investigation is needed to determine if the metrics performance of the proposed algorithm are the result of non-appropriate neural network architecture or the complexity of the connection between user attributes and written texts, in general.
However, the current method discovers a new approach for fake text review detection and can serve as a good baseline for future algorithms both for human biased reviews and generative language texts detection.

\section{Conclusions} \hyperlink{ch:5}{}
The current research discovers the novel approach towards fake text reviews detection in the collaborative filtering-based recommender systems.
The proposed method utilizes the user metadata, alongside the text reviews, as the additional evidence against fake reviews.
This way, our method, based on the contrastive learning setup, makes the recommender platform robust to two kinds of attack strategies: (1) the ones, generated by humans writing biased reviews for monetary gains, and (2) the ones, generated by language models.
The experimental results demonstrate the decent performance of the proposed algorithm on these two classification tasks, making it superior to the SOTA fakeRoBERTa \cite{salminen2022} method, which processes the texts only.
Moreover, we have significantly contributed by generating two data sets, which simulate both kinds of attacks.
The first data set is devoted to Honest Users vs Human Spammers classification task and is based on the Google Local Business Reviews and Metadata data set, which is further modified by engineering user demographic characteristics and is pre-processed via proposed sampling procedure.
The second data set is a collection of fake reviews, generated by GPT-J-6B \cite{gpt-j} model and augmented with random user metadata, which can be used for Honest Humans vs Generative Language Models classification task.
Overall, the current study proposes a more universal approach towards fake reviews detection, setting a good baseline for future research.

The current project has faced several research limitations.
First of all, the nature of the problem statement implies the usage of the data set, which contains both the review texts and the user demographic characteristics.
The available open-source recommender systems data sets contain either information about the text reviews or the user metadata, but not both.
The only open-source data set, which satisfies this condition, is the Google Local Business Reviews and Metadata, which we have conducted our numerical experiments on.
However, the experimental results that we have obtained are valid for this data set and may not be generalisable onto the other recommender system platforms.
Henceforth, there is the need to access the private industrial data sets with both review texts and user attributes to be at hand to validate our approach on.

Moreover, it has been noted that the problem statement is new for the recommender systems setup and has not been approached yet.
Therefore, there are no valid algorithms to compare the performance of our proposed approach with.
Henceforth, there is the need for further investigation into the matter to develop a more powerful algorithm, having the proposed one as a benchmark.

Another research limitation regards the connection between the review texts and the user demographic data.
We have assumed that the real users' metadata is latently connected to the text reviews, and that this connection can be grasped by the deep neural network.
It may be the case that the deep neural network architecture used is not very suitable for this kind of task to grasp this latent connection or that the connection itself is weak and cannot be grasped by any other model.
This limitation should be addressed more specifically through conducting the psychological and linguistic experiments and checking the strength of connection between users' demographic characteristics and the text reviews he/she writes.
Also, it may be useful to check the performance of tested approaches on other similar data sets, as defined above, to check how good the patterns are grasped along different data sets.

\begin{acks}
The authors thank Anton Voronov for productive discussion of the problem and for pointing out plausibility of using contrastive learning in the current setting.
\end{acks}

\bibliographystyle{ACM-Reference-Format}
\bibliography{main.bib}

\end{document}